\documentclass[conference]{IEEEtran}

\newcommand{\ourmaintitle}{Causal Discovery by Telling Apart Parents~and~Children}

\newcommand{\oururl}{\url{http://eda.mmci.uni-saarland.de/climb/}}

\usepackage{pgfplots}
\usepgfplotslibrary{fillbetween}
\usepackage{latexsym}
\usepackage{balance}
\usepackage{amsmath,graphicx,centernot}
\usepackage[ruled, vlined, nofillcomment, linesnumbered]{algorithm2e}
\usepackage[noenumitem]{eda} 
\usepackage{footmisc}
\usepackage{amsthm}

\graphicspath{ {./} }

\ifpdf
\usetikzlibrary{external}
\fi

\SetKwComment{tcpas}{\{}{\}}
\SetCommentSty{textnormal}
\SetArgSty{textnormal}
\SetKw{False}{false}
\SetKw{True}{true}
\SetKw{Null}{null}
\SetKwInOut{Output}{output}
\SetKwInOut{Input}{input}
\SetKw{AND}{and}
\SetKw{OR}{or}
\SetKw{Continue}{continue}

\newcommand{\climb}{\textsc{Climb}\xspace}

\newcommand{\ourmethod}{\climb}

\newcommand{\SC}{\mathit{SC}\xspace}
\newcommand{\SCI}{\ensuremath{\mathit{SCI}}\xspace}
\newcommand{\CI}{\ensuremath{I_{\SC}}\xspace}
\newcommand{\PC}{\ensuremath{\mathit{PC}}\xspace}

\newcommand{\pcmb}{\textsc{PCMB}\xspace}

\newcommand{\ipcmb}{\textsc{IPCMB}\xspace}
\newcommand{\findPC}{\textsc{FindPC}\xspace}
\newcommand{\PCalgo}{\textsc{PC}\xspace}
\newcommand{\fges}{\textsc{FGES}\xspace}
\newcommand{\bestPartition}{\textsc{FindBestPartition}\xspace}

\newcommand{\tMB}{MB\xspace}
\newcommand{\mMB}{\mathit{MB}\xspace}
\newcommand{\pa}{\mathit{PA}\xspace}
\newcommand{\ch}{\mathit{CH}\xspace}
\renewcommand{\sp}{\mathit{SP}\xspace}
\newcommand{\ci}{\mathit{ci}\xspace}

\newcommand{\nml}{\mathit{NML}\xspace}

\newcommand{\fallingfactorial}[1]{%
  ^{\mspace{2mu}\underline{\mspace{-2mu}#1\mspace{-2mu}}\mspace{2mu}}%
}
\newcommand{\risingfactorial}[1]{%
  ^{\mspace{2mu}\overline{\mspace{-2mu}#1\mspace{-2mu}}\mspace{2mu}}%
}

\newtheorem{postulate}{Postulate}
\newtheorem{proposition}{Proposition}
\newtheorem{definition}{Definition}
\newtheorem{lemma}{Lemma}

\renewcommand{\models}{\mathcal{M}\xspace}
\newcommand{\regret}{\mathcal{R}\xspace}

\newcommand\independent{\protect\mathpalette{\protect\independenT}{\perp}}
\def\independenT#1#2{\mathrel{\rlap{$#1#2$}\mkern2mu{#1#2}}}

	\tikzstyle{flatlabel}  = [above, font = \tiny, inner sep = 1pt, text = black]
	\tikzstyle{flatlabelb}  = [below, font = \tiny, inner sep = 1pt, text = black]
	\tikzstyle{slopelabel}  = [sloped, above, font = \tiny, inner sep = 1pt, text = black]
	\tikzstyle{slopelabelb}  = [sloped, below, font = \tiny, inner sep = 1pt, text = black]

\usepackage{tikz}
\usepackage{pgfplots}
\usepackage{pgfplotstable}
\usetikzlibrary{arrows,shapes,positioning,fit,calc,matrix,trees}
\usetikzlibrary{external}

\pgfdeclareplotmark{btriangle}{%
  \pgfpathmoveto{\pgfpoint{-\pgfplotmarksize}{\pgfplotmarksize}}
  \pgfpathlineto{\pgfpoint{0pt}{-\pgfplotmarksize}}
  \pgfpathlineto{\pgfpoint{\pgfplotmarksize}{\pgfplotmarksize}}
  \pgfpathlineto{\pgfpoint{-\pgfplotmarksize}{\pgfplotmarksize}}
  \pgfusepathqstroke
}
\pgfdeclareplotmark{btriangle*}{%
  \pgfpathmoveto{\pgfqpoint{0pt}{-\pgfplotmarksize}}
  \pgfpathlineto{\pgfqpointpolar{150}{\pgfplotmarksize}}
  \pgfpathlineto{\pgfqpointpolar{30}{\pgfplotmarksize}}
  \pgfpathclose
  \pgfusepathqfillstroke
}

\pgfdeclareplotmark{ltriangle}{%
  \pgfpathmoveto{\pgfqpoint{\pgfplotmarksize}{0pt}}
  \pgfpathlineto{\pgfqpointpolar{-120}{\pgfplotmarksize}}
  \pgfpathlineto{\pgfqpointpolar{120}{\pgfplotmarksize}}
  \pgfpathclose
  \pgfusepathqstroke
}
\pgfdeclareplotmark{ltriangle*}{%
  \pgfpathmoveto{\pgfqpoint{\pgfplotmarksize}{0pt}}
  \pgfpathlineto{\pgfqpointpolar{-120}{\pgfplotmarksize}}
  \pgfpathlineto{\pgfqpointpolar{120}{\pgfplotmarksize}}
  \pgfpathclose
  \pgfusepathqfillstroke
}

\pgfdeclareplotmark{rtriangle}{%
  \pgfpathmoveto{\pgfqpoint{-\pgfplotmarksize}{0pt}}
  \pgfpathlineto{\pgfqpointpolar{-60}{\pgfplotmarksize}}
  \pgfpathlineto{\pgfqpointpolar{60}{\pgfplotmarksize}}
  \pgfpathclose
  \pgfusepathqstroke
}

\definecolor{yafaxiscolor}{rgb}{0.3, 0.3, 0.3}
\definecolor{yafcolor1}{rgb}{0.4, 0.165, 0.553}
\definecolor{yafcolor2}{rgb}{0.949, 0.482, 0.216}
\definecolor{yafcolor3}{rgb}{0.47, 0.549, 0.306}
\definecolor{yafcolor4}{rgb}{0.925, 0.165, 0.224}
\definecolor{yafcolor5}{rgb}{0.141, 0.345, 0.643}
\definecolor{yafcolor6}{rgb}{0.965, 0.633, 0.267}
\definecolor{yafcolor7}{rgb}{0.627, 0.118, 0.165}
\definecolor{yafcolor8}{rgb}{0.878, 0.475, 0.686}

\pgfplotscreateplotcyclelist{yaf}{%
{yafcolor1,mark options={scale=0.45},mark=*},
{yafcolor5,mark options={scale=0.60},mark=triangle*},
{yafcolor3,mark options={scale=0.45},mark=square*},
{yafcolor4,mark options={scale=0.70},mark=btriangle*},
{yafcolor8,mark options={scale=0.60},mark=ltriangle*},
{yafcolor7,mark options={scale=0.60},mark=rtriangle*},
{yafcolor2,mark options={scale=0.60},mark=diamond*},
{yafcolor6,mark options={scale=0.60},mark=pentagon*}}

\tikzset{
precise pin/.style args={[#1][#2]#3:#4}{
    pin={[inner sep=0pt, #1, label={[append after command={
		node [#2,
			outer sep = 0pt,
			inner sep=0pt,
			at=(\tikzlastnode),
			anchor=#3+180 ] {#4} } ]center:{}}]#3:{}}
}}

\pgfplotsset{
	clip = false,
	clip marker paths = true,
	tick align=outside,
	x tick label style = {font=\scriptsize, yshift = 1pt},
	y tick label style = {font=\scriptsize, xshift = 1pt},
	major tick length = 2pt,
    every axis y label/.style = {at = {(ticklabel cs:0.5)}, rotate=90, anchor=center, font=\scriptsize, xshift = 2pt},
	every axis x label/.style = {at = {(ticklabel cs:0.5)}, anchor=center, font=\scriptsize, yshift = -2pt},
	axis y line*=left, axis x line*=bottom,
        enlargelimits = 0.03
}
\tikzstyle{every pin}=[font=\footnotesize, inner sep = 0pt, distance=2em]
\tikzstyle{every pin edge}=[line width = 0.1pt, pin distance = 2em]

\newlength{\myheight}
\newlength{\mywidth}

\newcommand{\legDist}{Distance ($\savg{\distc{}}$)}
\newcommand{\legJacc}{Jacc.\ dist.\ ($\savg{\jacc{}}$)}
\newcommand{\legCover}{Coverage ($\cover{}$)}
\newcommand{\legDens}{Density ($\savg{\density{}}$)}

\colorlet{graphcl1}{yafcolor1!50}
\colorlet{graphcl2}{yafcolor4!30}
\colorlet{graphcl3}{yafcolor2!50}
\colorlet{graphcl4}{yafcolor5}
\colorlet{graphcl5}{yafcolor4}
\colorlet{graphcl6}{yafcolor6}

\tikzstyle{graphedge} = [black, thick, opacity = 0.5]
\tikzstyle{graphnode} = [draw = black, circle, line width = 0pt, text = black, inner sep = 0.5pt, text width = 10pt, align = center]
\tikzstyle{outliernode} = [circle, line width = 0pt, draw, text = black, fill = white, inner sep = 0.5pt, text width = 10pt, align = center]

\tikzstyle{toyedge} = [->, black, thick, bend left = 10, yafcolor5]
\tikzstyle{toynode} = [draw = black, thick, circle, line width = 0pt, text = black, inner sep = 0pt, text width = 13pt, align = center]
\tikzstyle{groupline} = [black, thick, dashed]

\makeatletter
\tikzset{multicircle/.style  args={#1, (#2)}{%
 alias=tmp@name, %
  postaction={%
    insert path={
     \pgfextra{%
     \pgfpointdiff{\pgfpointanchor{\pgf@node@name}{center}}%
                  {\pgfpointanchor{\pgf@node@name}{east}}%
     \pgfmathsetmacro\insiderad{\pgf@x}%
     \foreach \c [count=\ci from = 0, evaluate=\ci as \angle using 360 - (\ci) * #1] in {#2}%
        \fill[\c] (\pgf@node@name.center)  -- ++(0:\insiderad-\pgflinewidth) arc (0:\angle:\insiderad-\pgflinewidth)--cycle;%
        }}}}}
\makeatother

\pgfplotsset{
  boxplot/box width/.initial=1em,
  solid boxes/.style={
    mark=x,
    boxplot/draw direction=y,
    boxplot/whisker extend=0,
    boxplot/draw/median/.code={%
      \draw[mark size=2pt,/pgfplots/boxplot/every median/.try]
        \pgfextra
        \pgftransformshift{
          \pgfplotsboxplotpointabbox
            {\pgfplotsboxplotvalue{median}}
            {0.5}
        }
        \pgfsetfillcolor{white}
        \pgfuseplotmark{*}
        \endpgfextra
      ;
    },
    boxplot/draw/box/.code={
      \draw[fill,/pgfplots/boxplot/every box/.try]
        ($(boxplot box cs:\pgfplotsboxplotvalue{lower quartile},0.5)!0.5\pgfkeysvalueof{/pgfplots/boxplot/box width}!(boxplot box cs:\pgfplotsboxplotvalue{lower quartile},0)$)
        rectangle
        ($(boxplot box cs:\pgfplotsboxplotvalue{upper quartile},0.5)!0.5\pgfkeysvalueof{/pgfplots/boxplot/box width}!(boxplot box cs:\pgfplotsboxplotvalue{upper quartile},1)$)
      ;
    }
  },
}

\begin{document}
\setlength{\pdfpagewidth}{8.5in}
\setlength{\pdfpageheight}{11in}

\title{\ourmaintitle}

\author{\IEEEauthorblockN{Alexander Marx and Jilles Vreeken}
\IEEEauthorblockA{Max Planck Institute for Informatics and Saarland University\\   
Saarland Informatics Campus, Saarbr\"{u}cken, Germany\\
\{amarx, jilles\}@mpi-inf.mpg.de}
}

\maketitle

\begin{abstract}
We consider the problem of inferring the directed, causal graph from observational data, assuming no hidden confounders. We take an information theoretic approach, and make three main contributions. 

First, we show how through algorithmic information theory we can obtain SCI, a highly robust, effective and computationally efficient test for conditional independence---and show it outperforms the state of the art when applied in constraint-based inference methods such as stable PC. 

Second, building upon on SCI, we show how to tell apart the parents and children of a given node based on the algorithmic Markov condition. We give the \ourmethod algorithm to efficiently discover the directed, causal Markov blanket---and show it is at least as accurate as inferring the global network, while being much more efficient. 

Last, but not least, we detail how we can use the \ourmethod score to direct those edges that state of the art causal discovery algorithms based on PC or GES leave undirected---and show this improves their precision, recall and F1 scores by up to 20\%. 
\end{abstract}

\begin{IEEEkeywords}
Conditional Independence, Stochastic Complexity, Markov Blankets, Causal Discovery
\end{IEEEkeywords}

\maketitle
\section{Introduction}
\label{sec:intro}

Many mechanisms, including gene regulation and control mechanisms of complex systems, can be modelled naturally by causal graphs. While in theory it is easy to infer causal directions if we can manipulate parts of the network---i.e. through controlled experiments---in practice, however, controlled experiments are often too expensive or simply impossible, which means we have to work with observational data~\cite{pearl:09:book}.

Constructing the causal graph given observations over its joint distribution can be understood as a global task, as finding the whole directed network, or locally, as discovering the local environment of a target variable in a causal graph~\cite{pearl:88:firstmb,spirtes:00:book}. For both problem settings, constraint based algorithms using conditional independence tests, belong to the state of the art~\cite{aliferis:03:hiton,fu:08:ipcmb,spirtes:00:book,pena:07:pcmb}.

As the name suggests, those algorithms strongly rely on one key component: the independence test. For discrete data, the state of the art methods often rely on the $G^2$ test~\cite{aliferis:10:hiton:overview,schluter:14:survey:mb}. While it performs well given enough data, as we will see, it has a very strong bias to indicating independence in case of sparsity. 
Another often used method is conditional mutual information (CMI)~\cite{zhang:10:iamb:lambda}, which like $G^2$ performs well in theory, but in practice has the opposite problem; in case of sparsity it tends to find spurious dependencies---i.e. it is likely to find no independence at all, unless we set arbitrary cut-offs.  

To overcome these limitations, we propose a new independence measure based on algorithmic conditional independence~\cite{janzing:10:algomarkov}, which we instantiate through the Minimum Description Length principle~\cite{grunwald:07:book,rissanen:78:mdl}. In particular, we do so using stochastic complexity~\cite{rissanen:86:stochastic}, which allows us to measure the complexity of a sample in a mini-max optimal way. That is, it performs as close to the true model as possible (with optimality guarantees), regardless of whether the true model lies \emph{inside} or \emph{outside} the model class~\cite{grunwald:07:book}. As we consider discrete data, we instantiate our test using stochastic complexity for multinomials. As we will show, our new measure overcomes the drawbacks of both $G^2$ and CMI, and performs much better in practice, especially under sparsity.

\begin{figure}[t]
	\centering
	\includegraphics[]{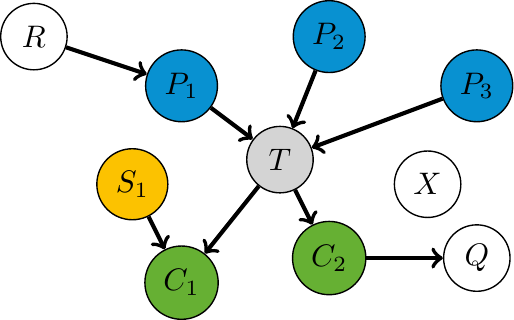}
	\caption{Example Markov blanket $MB_T$ of a target $T$, consisting of three parents (\textcolor{mambacolor4}{blue} nodes), two children (\textcolor{mambacolor3}{green}) and one spouse (\textcolor{mambacolor5}{orange}). All other nodes are $\independent T$ given $MB_T$. Our goal is to discover the MB, \emph{as well as} the edge directions.}
	\label{fig:sample_mb}
\end{figure}

Building upon our independence test, we consider the problem of finding the Markov blanket, short MB (see example in Fig.~\ref{fig:sample_mb}). Precisely, the Markov blanket of a target variable $T$ is defined as the minimal set of variables, conditioned on which all other variables are independent of the target~\cite{pearl:88:firstmb}. This set includes its parents, children and parents of common children, also called spouses. Simply put, the Markov blanket of a target node $T$ is considered as the minimal set of nodes that contains all information about $T$~\cite{aliferis:10:hiton:overview}. 

Algorithms for finding the Markov blanket stop at this point and return a set of nodes, without identifying the parents, the children or the spouses~\cite{aliferis:03:hiton,fu:08:ipcmb,pena:07:pcmb}. We propose \ourmethod, a new method based on the algorithmic Markov condition~\cite{janzing:10:algomarkov}, that is not only faster than state of the art algorithms for discovering the MB, but is to the best of our knowledge the first algorithm for discovering the directed, or \emph{causal} Markov blanket of a target node, without further exploration of the network. That is, it tells apart  parents, children and spouses.

Last but not least, we consider recovering the full causal graph. Current state of the art constraint based and score based algorithms~\cite{chickering:02:ges,spirtes:00:book} only discover partially directed graphs but can not distinguish between Markov equivalent sub-graphs. Based on \ourmethod, we propose a procedure to infer the remaining undirected edges with a high precision.

The key contributions of this paper are, that we
\begin{itemize}[noitemsep,topsep=0pt]
	\item[(a)] define \SCI, a new conditional independence test,
	\item[(b)] derive a score to tell apart parents and children of a node,
	\item[(c)] propose \ourmethod, to discover causal Markov blankets, and
	\item[(d)] show how to use the \ourmethod score to orient those edges that can not be oriented by the PC or GES algorithm.
\end{itemize}
This paper is structured as follows. In Sec.~\ref{sec:preliminaries} we introduce the main concepts and notation, as well as properties of the stochastic complexity. Sec.~\ref{sec:related} discusses related work. Since our contributions build upon each other, we first introduce \SCI, our new independence test, in Sec.~\ref{sec:independence}. Next, we define and explain the \climb algorithm to find the causal Markov blanket in Sec.~\ref{sec:climb} and extend this idea to decide between Markov equivalent DAGs. We empirically evaluate in Sec~\ref{sec:experiments} and round up with discussion and conclusion in Sec.~\ref{sec:conclusion}. 
\section{Preliminaries}
\label{sec:preliminaries}

In this section, we introduce the notation and main concepts we build upon in this paper. All logarithms are to base $2$, and we use the common convention that $0 \log 0 = 0$.

\subsection{Bayesian Networks}

Given an $m$-dimensional vector $(X_1, \dots, X_m)$, a Bayesian network defines the joint probability over random variables $X_1, \dots, X_m$. We specifically consider discrete data, i.e. each variable $X_i$ has a domain $\mathcal{X}_i$ of $k_i$ distinct values. Further, we assume we are given $n$ i.i.d. samples drawn from the joint distribution of the network. We express the $n$-dimensional data vector for the $i$-th node with $x_i^n$. To describe interactions between the nodes, we use a directed acyclic graph (DAG) $G$. We denote the parent set of a node $X_i$ with $\pa_i$, its children with $\ch_i$ and nodes that have a common child with $X_i$ as its spouses $\sp_i$. A set of variables $\mathbf{X}$ contains $k_{\mathbf{X}} = \prod_{X_j \in \mathbf{X}} k_j$ value combinations that can be non-ambiguously enumerated. We write $\mathbf{X} = j$ to refer to the $j$-th value combination of $\mathbf{X}$. For instance such a set could be the set of parents, children or spouses of a node.

As it is common for both inferring the Markov blanket as well as the complete network, we assume causal sufficiency, that is, we assume that we have measured all common causes of the measured variables. Further, we assume that the Bayesian network $G$ is faithful to the underlying probability distribution $P$~\cite{spirtes:00:book}. 

\begin{figure}[t]
	\centering
	\includegraphics[]{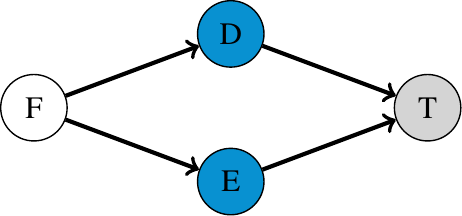}
	\caption{[D-Separation] Given the above causal DAG it holds that $F \independent T \mid D,E$, or $F$ is d-separated of $T$ given $D, E$. Note that $D \not\independent T \mid E,F$ and $E \not\independent T \mid D,F$.}
	\label{fig:d_separation}
\end{figure}

\begin{definition}[Faithfulness]
If a Bayesian network $G$ is faithful to a probability distribution $P$, then for each pair of nodes $X_i$ and $X_j$ in $G$, $X_i$ and $X_j$ are adjacent in $G$ iff. $X_i \not \independent X_j \mid \textbf{Z}$, for each $\textbf{Z} \subset G$, with $X_i, X_j \not \in \textbf{Z}$.
\end{definition}

Equivalently, it holds that $X_i$ and $X_j$ are $d$-separated by $\textbf{Z}$, with $\textbf{Z} \subset G$ and $X_i, X_j \not \in \textbf{Z}$, iff. $X_i \independent X_j \mid \textbf{Z}$~\cite{janzing:10:algomarkov}. Generally, $d$-separation is an important concept for constraint based algorithms, since it is used to prune out false positives. As an example consider Fig.~\ref{fig:d_separation}. All nodes $D,E$ and $F$ are associated with the target $T$. However, $F \independent T \mid D,E$ and hence $F$ is $d$-separated from $T$ given $D$ and $E$, which means that it can be excluded from the parent set of $T$.

Building on the faithfulness assumption, it follows that the probability to describe the whole network can be written as
\begin{equation}
P(X_1, \dots, X_m) = \prod_{i = 1}^m P(X_i \mid \pa_i) \; , \label{eq:algmarkov:network}
\end{equation}
which means that we only need to know the conditional distributions for each node $X_i$ given its parents~\cite{janzing:10:algomarkov}. 

Having defined a Bayesian network, it is now easy to explain what a Markov blanket is.

\subsection{Markov Blankets}

Markov blankets were first described by Pearl~\cite{pearl:88:firstmb}. A Markov blanket $\mMB_T$ of a target $T$ is the minimal set of nodes in a graph $G$, given which all other nodes are conditionally independent of $T$. That is, knowing the values of $\mMB_T$, we can fully explain the probability distribution of $T$ and any further information is redundant~\cite{pearl:88:firstmb}. Concretely, the Markov blanket of $T$ consists of the parents, children and spouses of $T$.  An example \tMB is shown in Fig.~\ref{fig:sample_mb}. 

Discovering the Markov blanket of a node has several advantages. First, the Markov blanket contains all information that we need to describe a target variable as well as its neighbourhood in the graph. In addition, the \tMB is theoretically the optimal set of attributes to predict the target values~\cite{margaritis:00:gs} and can be used for feature selection~\cite{aliferis:10:hiton:overview,fu:10:review:mb:featureselection}. In addition to those properties, the Markov blanket of a single target can be inferred much more efficiently than the whole Bayesian network~\cite{visweswaran:14:mbcounting}. This is especially beneficial if we are only interested in a single target in a large network, e.g. the activation of one gene.

Most algorithms to discover the \tMB rely on faithfulness and a conditional independence test~\cite{aliferis:10:hiton:overview}. For the former, we have to trust the data, while for the latter we have a choice. To introduce the independence test we propose, we first need to explain the notions of Kolmogorov complexity and Stochastic complexity.

\subsection{Kolmogorov Complexity}

The Kolmogorov complexity of a finite binary string $x$ is the length of the shortest binary program $p^*$ for a universal Turing machine $\mathcal{U}$ that generates $x$, and then halts~\cite{kolmogorov:65:information, vitanyi:93:book}. Formally, we have
\[
K(x) = \min \{ |p| \mid p \in \{0,1\}^*, \mathcal{U}(p) = x \} \; .
\]
That is, program $p^*$ is the most succinct \emph{algorithmic} description of $x$, or in other words, the ultimate lossless compressor for that string. We will also need conditional Kolmogorov complexity, $K(x \mid y) \leq K(x)$, which is again the length of the shortest binary program $p^*$ that generates $x$, and halts, but now given $y$ as input for free. 

By definition, Kolmogorov complexity will make maximal use of any structure in $x$ that can be expressed more succinctly algorithmically than by printing it verbatim. As such it is the theoretical optimal measure for complexity. However, due to the halting problem it is not computable, nor approximable up to arbitrary precision~\cite{vitanyi:93:book}. We can, however, approximate it from above through MDL.

\subsection{The Minimum Description Length Principle}
 
The Minimum Description Length (MDL) principle~\cite{rissanen:78:mdl} provides a statistically well-founded and computable framework to approximate Kolmogorov complexity from above~\cite{grunwald:07:book}. In refined MDL we measure the \emph{stochastic complexity} of data $x$ with regard to a model class $\models$, $L(x \mid \models) \geq K(x)$. The larger this model class, the closer we get to $K(x)$; the ideal version of MDL considers the set of \emph{all} programs that we \emph{know} output $x$ and halt, and hence is the same as Kolmogorov complexity. 
By using a \textit{refined} MDL code $\bar{L}$, we have the guarantee that $\bar{L}(x\mid\models)$ is only a constant away from the number of bits we would need to encode the data if we already \emph{knew} the best model 
This constant is called the \emph{regret}. Importantly, it does not depend on the data, but only on the model class; and hence these guarantees hold even in settings where the data was drawn adversarially~\cite{grunwald:07:book}. 
Only for a few model classes it is known how to efficiently compute the stochastic complexity. One of these is the stochastic complexity for multinomials~\cite{kontkanen:07:histo}, which we will introduce below.

\subsection{Stochastic Complexity for Multinomials}

Given $n$ samples of a discrete-valued univariate attribute $X$ with a domain $\mathcal{X}$ of $k$ distinct values, $x^n \in
\mathcal{X}^n$, 
let $\hat{\theta}(x^n)$ denote the maximum likelihood estimator for $x^n$. Shtarkov~\cite{shtarkov:87:universal} defined the mini-max optimal \textit{normalized maximum likelihood (NML)} as
\begin{equation}
P_{\nml}(x^n \mid \models_k) = \frac{P(x^n \mid \hat{\theta}(x^n, \models_k))}{\regret_{\models_k}^n} \; , \label{eq:pnml}
\end{equation}
where the normalizing factor, or regret $\regret_{\models_k}^n$, relative to the model class $\models_k$ is defined as
\begin{equation}
\regret_{\models_k}^n = \sum_{x^n \in \mathcal{X}^n} P(x^n \mid \hat{\theta}(x^n), \models_k) \, . \label{eq:regret}
\end{equation}
The sum goes over every possible $x^n$ over the domain of $X$, and for each considers the maximum likelihood for that data given model class $\models_k$. For discrete data, we can rewrite Eq.~\eqref{eq:pnml} as
\[
P_{\nml}(x^n \mid \models_k) = \frac{\prod_{v=1}^k \left(\frac{h_v}{n} \right)^{h_v}}{\regret_{\models_k}^n} \; ,
\]
writing $h_v$ for the frequency of value $v$ in $x^n$, resp.\ Eq.~\eqref{eq:regret} as
\[
\regret_{\models_k}^n = \sum_{h_1 + \cdots + h_k = n} \frac{n!}{h_1! \cdots h_k!} \prod_{v=1}^k \left( \frac{h_v}{n} \right)^{h_v} \ .
\]
Mononen and Myllym{\"{a}}ki~\cite{mononen:08:sub-lin-stoch-comp} derived a formula to calculate the regret in sub-linear time, meaning that the whole formula can be computed in linear time w.r.t. the number of samples $n$.

We obtain the stochastic complexity for $x^n$ with regard to model class $\models_k$
by simply taking the negative logarithm of the $P_{\nml}$, 
which decomposes into a Shannon-entropy and a log regret term,
\begin{align}
SC(x^n \mid \models_k) &= - \log P_{\nml}(x^n \mid \models_k) \; ,\\
&= n H(x^n) + \log \regret_{\models_k}^n \; . \label{eq:sc:unc}
\end{align}
Conditional stochastic complexity~\cite{silander:08:nml:bayesnet} is defined analogue to conditional entropy, i.e. we have for any $x^n$, $y^n$ that
\begin{align}
SC(x^n \mid y^n, \models_k) &= \sum_{v \in \mathcal{Y}} SC(x^n \mid y^n = v, \models_k) \\
&= \sum_{v \in \mathcal{Y}} h_v H(x^n \mid y^n = v) + \sum_{v\in \mathcal{Y}} \log \regret_{\models_k}^{h_v}\; ,\label{eq:sc:cond}
\end{align}
with $\mathcal{Y}$ the domain of $Y$, and $h_v$ is the frequency of a value $v$ in $y^n$. 

For notational convenience, wherever clear from context we will write $SC(X)$ for $SC(x^n \mid \models_k)$, resp. $SC(X \mid Y)$ for $SC(x^n \mid y^n, \models_k)$. For the log-regret terms in Eq.~\eqref{eq:sc:unc}, resp. Eq.~\eqref{eq:sc:cond}, we will write $\Delta(X)$ resp. $\Delta(X \mid \cdot)$. To introduce our independence test and methods, we need the following property of the conditional stochastic complexity.

\begin{proposition}
\label{prop:monotone}
Given two discrete random variables $X$ and $Y$ and a set $\mathbf{Z}$ of discrete random variables. If $\Delta(X \mid \mathbf{Z}) \le \Delta(X \mid \mathbf{Z},Y)$, we call $\Delta$ monotone w.r.t. the number of conditioning variables.
\end{proposition}

To proof Prop.~\ref{prop:monotone}, we first show that the regret term is log-concave in $n$. 

\begin{lemma}
\label{lemma:log:concave}
For $n \ge 1$, the regret term $\regret_{\models_k}^n$ of the multinomial stochastic complexity of a random variable with a domain size of $k \ge 2$ is log-concave in $n$.
\end{lemma}

For consciousness, we postpone the proof of Lemma~\ref{lemma:log:concave} to Appendix~\ref{app:concave}. Based on Lemma~\ref{lemma:log:concave} we can now proof Prop.~\ref{prop:monotone}.

\begin{IEEEproof}[Proof of Prop.~\ref{prop:monotone}]
In order to show Prop.~\ref{prop:monotone}, we can reduce the statement as follows. Consider that $\mathbf{Z}$ contains $p$ distinct value combinations $\{ r_1, \dots, r_p \}$. If we add $Y$ to $\mathbf{Z}$, the number of distinct value combinations, $\{ l_1, \dots, \l_q \}$, increases to $q$, where $p \le q$. Consequently, to show that Prop.~\ref{prop:monotone} holds, it suffices to show that
\begin{equation}
\sum_{i = 1}^p \log \regret_{\models_k}^{|r_i|} \le \sum_{j = 1}^q \log \regret_{\models_k}^{|l_j|} \, \label{eq:sub:additivity:preparation}
\end{equation}
whereas $\sum_{i=1}^p |r_i| = \sum_{j=1}^q |l_j| = n$. Next, consider w.l.o.g. that each value combination $\{r_i\}_{i=1, \dots, p}$ is mapped to one or more value combinations in $\{ l_1, \dots, \l_q \}$. Hence, Eq.~\eqref{eq:sub:additivity:preparation} holds, if the $\log \regret_{\models_k}^n$ is sub-additive in $n$. Since we know from Lemma~\ref{lemma:log:concave} that the regret term is log-concave in $n$, sub-additivity follows by definition.
\end{IEEEproof}
\section{Related Work}
\label{sec:related}

Many methods for discovering Markov blankets and causal networks rely on conditional independence tests~\cite{aliferis:10:hiton:overview,spirtes:00:book}---which, by conditioning on the empty set, can also be used to measure association. For discrete data, the state of the art commonly relies on either the $G^2$ test or conditional mutual information (CMI)~\cite{aliferis:10:hiton:overview,fu:10:review:mb:featureselection,zhang:10:iamb:lambda}. Both measures, however, have drawbacks. The $G^2$ test is biased to independence when only limited data is available, whereas CMI has the opposite problem. In particular, it holds that $H(X \mid Y) \ge H(X \mid Y, Z)$ for any $X, Y, Z$, even if $Z \independent  X$. For both $G^2$ and CMI it is necessary to define a significance threshold or cut-off, which can be arbitrary. Our proposed method based on algorithmic independence overcomes these limitations.

The discovery of Markov blankets is important in two regards. First, it represents the optimal set of variables for feature selection~\cite{aliferis:10:hiton:overview} and second, for investigating the local structure around a target it is much faster than discovering the whole Bayesian network~\cite{visweswaran:14:mbcounting}. The idea of first discovering the neighbourhood of a node, instead of the full Bayesian network got more common with the grow and shrink (GS) algorithm~\cite{margaritis:00:gs}. It consists of two sub routines. First, it discovers the potential parents and children in a bottom up approach. Then, it finds the spouses based on the parents and children from the previous step.

To the best of our knowledge, there exists no algorithm that directly discovers the directed, causal Markov blanket, i.e. that can tell apart parents, children and spouses given only the Markov blanket of a target. In \climb, we first discover the Markov blanket, and then orient the edges. To discover the blanket, we build upon and extend the state of the art \pcmb~\cite{pena:07:pcmb} and \ipcmb~\cite{fu:08:ipcmb} algorithms. Both follow the general structure of the GS algorithm, with \ipcmb employing fast symmetry correction to exclude children of children.
Zhu and Yang~\cite{zhu:14:fast:ipcmb} proposed to speed up \ipcmb by pre-filtering based on mutual information, whereas Zhang et al.~\cite{zhang:10:iamb:lambda} discover Markov blankets based on conditional mutual information. Unlike our approach, these approaches require the user to set a cut-off value and a scaling parameter $\lambda$. 
For an in depth overview of related algorithms, we refer to the following articles~\cite{aliferis:10:hiton:overview,schluter:14:survey:mb}. 

In the area of causal discovery, constraint based~\cite{spirtes:00:book} and search and score based methods~\cite{glymour:99:computation}, are most well known. The state of the art constraint based algorithm is the Peter and Clark (PC) algorithm~\cite{spirtes:00:book}. It first discovers the skeleton of the causal graph, using conditional independence tests and then applies orientation rules to find the edge directions. 
There exist several extension to it, from which stable PC~\cite{colombo:14:stablepc}, which is independent of the order of the variables, has shown to perform well in practice. The second class of algorithms that we consider, are search and score based methods as the Greedy Equivalence Search (GES) algorithm~\cite{chickering:02:ges} and its improved version fast GES (\fges)~\cite{ramsey:17:fges}. Those algorithms are accurate even for small sample sizes and use a two stage greedy heuristic to discover the causal skeleton, which makes them very fast in practice.

Still the PC algorithm and \fges only infer partial DAGs and cannot decide between Markov equivalent sub-structures. Based on \ourmethod and the algorithmic Markov condition, we propose a post-processing step to infer the direction of those edges that cannot be oriented by stable PC and \fges.
\section{Independence Testing using Stochastic Complexity}
\label{sec:independence}

Most algorithms to discover the Markov blanket rely on two tests: association and conditional independence~\cite{aliferis:10:hiton:overview,fu:10:review:mb:featureselection}. 
The association test has to be precise for two reasons. First, by being too restrictive it might miss dependencies, which results in a bad recall. 
Second, if the test is too lenient we have to test more candidates. This may not sound bad, but as we face an exponential runtime w.r.t. the number of candidates, it is very much so in practice.

The quality of the conditional independence test is even more important. Algorithms proven to be correct, are only correct under the assumption that the conditional independence test is so, too~\cite{fu:10:review:mb:featureselection,pena:07:pcmb}. Commonly used conditional independence tests for categorical data like the $G^2$ or the conditional mutual information (CMI), have good properties in the limit, but show several drawbacks on practical sample sizes. In the following, we will propose and justify a new test for conditional independence on discrete data.

We start with Shannon conditional mutual information as a measure of independence. It is defined as follows.
\begin{definition}[Shannon Conditional Independence~\cite{cover:06:elements}]
Given random variables $X,Y$ and $\mathbf{Z}$. If
\begin{equation}
I(X ; Y \mid \mathbf{Z}) = H(X \mid \mathbf{Z}) - H(X \mid \mathbf{Z},Y) = 0
\label{eq:shannonindep}
\end{equation}
then $X$ and $Y$ are called statistically independent given $\mathbf{Z}$. 
\end{definition}

In essence, $I(X;Y \mid \mathbf{Z})$ is a measure of association of $X$ and $Y$ conditioned on $\mathbf{Z}$, where an association of $0$ corresponds to statistical independence. If $\mathbf{Z}$ is the empty set it reduces to standard mutual information, meaning we directly measure the association between $X$ and $Y$.

As $I$ is based on Shannon entropy, it assumes that we know the true distributions. In practice, we of course do not know these, but rather estimate $\hat{H}$ from finite samples. This becomes a problem when estimating conditional entropy, as to obtain a reasonable estimate we need a number of samples that is exponential in the domain size of the conditioning variable~\cite{mandros:17:noname}; if we have too few samples, we tend to underestimate the conditional entropy, overestimate the conditional mutual information, and Eq.~\eqref{eq:shannonindep} will seldom be $0$---even when $X \independent Y \mid \mathbf{Z}$. 

Hence, we needed to set an arbitrary cut-off $\delta$, such that $I \le \delta$. The problem is, however, that $\delta$ is hard to define, since it is dependent on the complexity of the variables, the amount of noise and the sample size.

Therefore, we propose a new independence test based on (conditional) algorithmic independence that remains zero for $X \independent Y \mid \mathbf{Z}$ even in settings where the sample size is small and the complexity of the variables is high. This we can achieve, by not only considering the complexity of the data under the model (i.e. the entropy) but also the complexity of the model (i.e. distribution). Before introducing our test, we first need to define algorithmic conditional independence.

\begin{definition}[Algorithmic Conditional Independence]
Given the strings $x,y$ and $z$, We 
write $z^*$ to denote the shortest program for $z$, and analogously $(z,y)^*$ for the shortest program for the concatenation of $z$ and $y$. If
\begin{equation}
I_A(x ; y \mid z) := K(x \mid z^*) - K(x \mid (z,y)^*) \stackrel{+}{=} 0 
\label{eq:algorithmicindep}
\end{equation}
holds up to an additive constant that is independent of the data, then $x$ and $y$ are called algorithmically independent given $z$~\cite{chaitin:75:algindepcond}. 
\end{definition}
As discussed above, Kolmogorov complexity is not computable, and to use $I_A$ in practice we will need to instantiate it through MDL. We do so using stochastic complexity for multinomials. That is, we rewrite Eq.~\eqref{eq:algorithmicindep} in terms of stochastic complexity, 
\begin{align}
I_{\SC}(X;Y \mid \mathbf{Z}) &= \SC(X \mid \mathbf{Z}) - \SC(X \mid \mathbf{Z},Y) \label{eq:scih} \\
 &= n \cdot I(X;Y \mid \mathbf{Z}) + \Delta(X \mid \mathbf{Z}) - \Delta(X \mid \mathbf{Z},Y)
\end{align}
where $n$ is the number of samples. Note that the regret terms $\Delta(X \mid \mathbf{Z})$ and $\Delta(X \mid \mathbf{Z},Y)$ in Eq.~\eqref{eq:scih} are over the same domain. From Prop~\ref{prop:monotone}, we know that $\Delta(X \mid \mathbf{Z})$ is smaller or equal than $\Delta(X \mid \mathbf{Z},Y)$. Hence, the new variable $Y$ has to provide a significant gain in the term $H(X \mid \mathbf{Z},Y)$ in Eq.~\eqref{eq:shannonindep} to overcome the penalty from its regret term. 

To use $I_{\SC}$ as an independence measure, we need one further adjustment. Since the regret terms are dependent on the domain size, it can happen that $I_{\SC}(X;Y \mid \mathbf{Z}) \neq I_{\SC}(Y;X \mid \mathbf{Z})$. We make the score symmetric by simply taking the maximum of both directions, and define the \textit{Stochastic Complexity based Independence} measure as
\begin{equation}
\SCI(X;Y \mid \mathbf{Z}) = \max \{ I_{\SC}(X;Y \mid \mathbf{Z}), I_{\SC}(Y;X \mid \mathbf{Z}) \} \label{eq:sci} \; .
\end{equation}
We have that $X \independent Y \mid \mathbf{Z}$, iff $\SCI(X;Y \mid \mathbf{Z}) \le 0$. Note that $\SCI$ can be smaller than zero, if e.g. $H(X \mid \mathbf{Z}) = H(X \mid Y, \mathbf{Z})$ but $\Delta(X \mid \mathbf{Z}) < \Delta(X \mid \mathbf{Z},Y)$. 

In the following, we explain why the \SCI is a well defined measure for conditional independence. In particular, we show that it detects independence, i.e. $\SCI(X;Y \mid \mathbf{Z}) \le 0$ holds, if $X \independent Y \mid \mathbf{Z}$ and that it converges to $I$.

\begin{lemma}
\label{lemma:sci_indep}
$\SCI(X;Y\mid \mathbf{Z}) \le 0$, iff $X \independent Y \mid \mathbf{Z}$.
\end{lemma}

\begin{IEEEproof}[Proof of Lemma \ref{lemma:sci_indep}]
It suffices to show that $\CI(X;Y \mid \mathbf{Z}) \le 0$, as $\CI(Y;X \mid \mathbf{Z}) \le 0$ follows analogously. Since the first part of $\CI(X;Y \mid \mathbf{Z})$ is equal to $n$ times $I$, this part will be zero by definition. Based on Prop~\ref{prop:monotone}, we have that $\Delta(X \mid \mathbf{Z}) - \Delta(X \mid \mathbf{Z},Y) \le 0$, which concludes the argument.
\end{IEEEproof}

Next, we show that in the limit $\frac{1}{n}\SCI(X;Y \mid \mathbf{Z})$ behaves like $I(X;Y \mid \mathbf{Z})$.

\begin{lemma}
\label{lemma:scii}
Given two random variables $X$ and $Y$ and a set of random variables $\mathbf{Z}$, it holds that 
\[
\lim_{n \rightarrow \infty} \frac{1}{n} \SCI(X; Y \mid \mathbf{Z}) = I(X; Y \mid \mathbf{Z}) \; ,
\]
whereas $n$ denotes the number of samples.
\end{lemma}
\begin{IEEEproof}[Proof of Lemma \ref{lemma:scii}]
To show the claim, it suffices to show that $\CI(X;Y \mid \mathbf{Z})$ asymptotically behaves like $I(X;Y \mid \mathbf{Z})$, as $\CI(Y;X \mid \mathbf{Z})$ has the same asymptotic behaviour. We have
\begin{align}
&\lim_{n \rightarrow \infty} \frac{1}{n} \CI(X;Y \mid \mathbf{Z}) \\
= &\lim_{n \rightarrow \infty} I(X;Y \mid \mathbf{Z}) + \frac{1}{n} \left( \Delta(X \mid \mathbf{Z}) - \Delta(X \mid \mathbf{Z}, Y) \right) \; .
\end{align}
Hence it remains to show that the second term goes to zero. Since $\log \regret_{\models_k}^n$ is concave in $n$, $\frac{1}{n} \Delta(X \mid \mathbf{Z})$ and $\frac{1}{n} \Delta(X \mid \mathbf{Z}, Y)$ will approach zero if $n \rightarrow \infty$.
\end{IEEEproof} 

In sum, asymptotically \SCI behaves like conditional mutual information, but in contrast to $I$, it is robust given only few samples, and hence does not need an arbitrary threshold. In practice it also performs favourably compared to the $G^2$ test, as we will show in the experiments.

Next, we build upon \SCI and introduce \ourmethod for discovering causal Markov blankets.
\section{Causal Markov Blankets}
\label{sec:climb}

In this section, we introduce \ourmethod, to discover directed, or causal Markov blankets.
As an example, consider Fig.~\ref{fig:sample_mb} again and further assume that we only observe $T$, its parents $P_1, P_2, P_3$ and its children $C_1, C_2$. Only in specific cases we can identify some of the parents using conditional independence tests. In particular, only where exist at least two parents $P_i$ and $P_j$ that are not connected by an edge or do not have any ancestor in common, i.e. when $P_i \independent P_j \mid \emptyset$ but $P_i \not \independent P_j \mid T$. We hence need another approach
to tell all apart the parents and children of $T$.

\subsection{Telling apart Parents and Children}
To tell apart parents and children, we define a partition $\pi(\PC_T)$ on the set of parents and children of a target node $T$, which separates the parents and children into exactly two non-intersecting sets. We refer to the first set as the parents $\pa_T$ and to the second as children $\ch_T$, for which $\pa_T \cup \ch_T = \PC_T$ and $\pa_T \cap \ch_T = \emptyset$. Further, we consider two special cases, where we allow either $\PC_T$ or $\ch_T$ to be empty, which leaves the remaining set to contain all elements of $\PC_T$. Note that there exist $2^{|\PC_T|}-1$ possible partitions of $\PC_T$. To decide which of the partitions fits best to the given data, we need to be able to score a partition.

Building on the faithfulness assumption, we know that we can describe the each node as the conditional distribution given its parents (see Eq.~\refeq{eq:algmarkov:network}). For causal networks, Janzing and Sch{\"o}lkopf~\cite{janzing:10:algomarkov} showed this equation can be expressed in terms of Kolmogorov complexity. 

\begin{postulate}[Algorithmic Independence of Conditionals] 
\label{postulate:algmarkov}
A causal hypothesis is only acceptable if the shortest description of the joint density $P$ is given by the concatenation of the shortest description of the Markov kernels. Formally, we write
\begin{equation}
K(P(X_1,\dots,X_m)) \stackrel{+}{=} \sum_j K(P(X_j \mid \pa_j)) \; , \label{eq:markov}
\end{equation}
which holds up to an additive constant independent of the input, and where $\pa_j$ corresponds to the parents of $X_j$ in a causal directed acyclic graph (DAG).
\end{postulate}

Further, Janzing and Sch{\"o}lkopf~\cite{janzing:10:algomarkov} show that this equation only holds for the true causal model. This means that it is minimal if each parent is correctly assigned to its corresponding children. Like in \SCI, we again use stochastic complexity to approximate Kolmogorov complexity. In particular, we reformulate Eq.~\refeq{eq:markov}, such that we are able to describe the local neighbourhood of a target node $T$ by its parents and children. In other words, we score a partition $\pi$ as 
\begin{align}
\SC(\pi(\PC_T)) &= \SC(T \mid \pa_T) + \sum_{P \in \pa_T} \SC(P) \\
&+ \sum_{C \in \ch_T} \SC(C \mid T) \label{eq:score:partition} \; .
\end{align}
where, we calculate the costs of $T$ given its parents, the unconditioned costs of the parents and the children given $T$. Further, by MDL, the \emph{best partition} $\pi^*(\PC_T)$ is the one minimizing Eq.~\eqref{eq:score:partition}. By exploring the whole search space, we can find the optimal partition with regard to our score. The corresponding computational complexity, which is exponential in the number of parents and children, is not the bottle neck for finding the causal Markov blanket, as it does not exceed the runtime for finding the parents and children in the first place. Moreover, in most real world data sets the average number of parents and children is rather small, leaving us on average with few computational steps here.

\subsection{The Climb Algorithm}

Now that we defined how to score a partition, we can now introduce the \ourmethod algorithm. In essence, the algorithm builds upon and extends \pcmb and \ipcmb, but, unlike these, can discover the causal Markov blanket.

\ourmethod (Algorithm~\ref{alg:climb}) consists of three main steps: First, we need to find the parents and children of the target node $T$ (line \ref{alg:findPC}), which can be done with any sound parents and children algorithm. Second, we compute the best partition $\pi^*(\PC_T)$ using Eq.~\eqref{eq:score:partition} (line \ref{alg:bestPartition}). The last step is the search for spouses. Here, we only have to iterate over the children to find spouses, which saves computational time. 
To remove children of children, we apply the fast symmetry correction criterion as suggested by Fu et al.~\cite{fu:08:ipcmb} (lines~\ref{alg:symmetryCorrection}--\ref{alg:continue}). Further, we find the spouses as suggested in the \pcmb algorithm~\cite{pena:07:pcmb} (lines~\ref{alg:searchForSpouse}--\ref{alg:addSpouse}), whereas the separating set $\mathbf{S}$ (line~\ref{alg:sepset}) can be recovered from the procedure that found the parents and children. In the last line, we output the causal MB by returning the distinct sets of parents, children and spouses.

\begin{algorithm}[tb!]
	\caption{\ourmethod}
	\label{alg:climb}
	\Input{data set $D$, target node $T$}
	\Output{the causal MB of $T$}
	$\PC = \findPC(T, D)$\; \label{alg:findPC}
	$\pa, \ch = \bestPartition(T, \PC)$\; \label{alg:bestPartition}
	$\sp = \emptyset$\;
	\ForEach{$C \in \ch$}{
		$\PC_C = \findPC(C, D)$\;
		\If{$T \notin \PC_C$}{ \label{alg:symmetryCorrection}
			$\ch = \ch \backslash \{ C \}$\; \label{alg:removeFP}
			continue\; \label{alg:continue}
		}
		\ForEach{$Y \in \PC_C \text{ and } Y \notin \{\pa, \ch, \sp \}$}{ \label{alg:searchForSpouse}
			$\mathbf{S} \subseteq \PC: T \independent Y \mid \mathbf{S}$\; \label{alg:sepset}
			\If{$T \not \independent Y \mid \mathbf{S} \cup \{ C \}$}{ \label{alg:checkSpouse}
				$\sp = \sp \cup \{ Y \}$\; \label{alg:addSpouse}
			}
		}
	}
	\Return{$\pa, \ch, \sp$}\;
\end{algorithm}

\subsubsection*{Complexity and Correctness}

At worst, the computational complexity of \ourmethod is as good as common Markov blanket discovery algorithms. This worst case is the scenario of having only children and therefore having to search each element in the parents and children set for the spouses. Given $|V|$ as the number of nodes, we have to apply $\mathcal{O}(2^{|MB|}|V|)$ independence tests, which reduces to $\mathcal{O}(|MB|^k|V|)$, if we restrict the number of conditioning variables in the independence test to $k$~\cite{aliferis:03:hiton}. Calculating the independence test is linear, as calculating the the conditional mutual information takes linear time and the regret term of \SCI can be computed in sub-linear time~\cite{mononen:08:sub-lin-stoch-comp}. In practice, \ourmethod saves a lot of computation compared to \pcmb or \ipcmb because it can identify the children and hence does not need to iterate through the parents to search for spouses.

If we search through the whole set of parents and children to identify the spouses, the correctness of the Markov blanket under the faithfulness condition and a correct independence test follows trivially from \pcmb~\cite{pena:07:pcmb} and \ipcmb~\cite{fu:08:ipcmb}. To correctly infer the \emph{causal} Markov blanket, we need to minimize our score over the complete causal network, which scales exponentially and is infeasible for networks with more than $33$ nodes~\cite{silander:08:nml:bayesnet}. Additionally, we would need the skeleton of the causal network, which would eliminate the computational advantage of only discovering the Markov blanket. Instead, we compute the local optimal score to tell apart parents and children. This makes \ourmethod feasible for large networks.
\subsection{Deciding between Markov equivalent DAGs}

In the previous section, we showed how to find the causal MB by telling apart parents and children. Besides, we can use this information to enhance current state of the art causal discovery algorithms. In particular, many causal discovery algorithms as GES~\cite{chickering:02:ges} and the \PCalgo~\cite{spirtes:00:book} algorithm find partial DAGs. That is, they can not orient all edges and leave some of them undirected. Precisely, if we would assign any direction at random to these undirected edges, the corresponding graph would be in the same Markov equivalence class as the original partial DAG.

We can, however, use the score of \ourmethod to also orient these edges in a post processing step as follows. First, we determine the parents and children of each node using the partial DAG. For an undirected edge connecting two nodes $A$ and $B$, we assign $B$ as a parent of $A$ and vice versa $A$ as a parent of $B$. It is easy to see that such an assignment creates a loop between $A$ and $B$ in the causal graph. In the second step, we iteratively resolve loops between two nodes $A$ and $B$ by we determining that configuration with the minimum costs according to Eq.~\ref{eq:score:partition}. This we do by first assigning $B$ as a parent of $A$ in $PC_A$, while keeping $A$ as a child of $B$ in $PC_B$ . We compute the sum of the costs of this configuration for $A$ and $B$ according to Eq.~\ref{eq:score:partition}, compare the result to the costs of the inverse assignment and select the one with smaller costs. We repeat this until all edges have been directed.

In the experiments we show that this simple technique significantly improves the precision, recall, and F1 in edge directions for both stable \PCalgo and \fges.
\section{Experiments}
\label{sec:experiments}

In this section, we empirically evaluate our independence test, as well as the \climb algorithm and the corresponding edge orientation scheme. For research purposes, we provide the code for \SCI, \ourmethod and the orientation scheme online.\!\footnote{\oururl\label{fn:apendix}} All experiments were ran single threaded on a Linux server with Intel Xenon E5-2643v2 processors and 64GB RAM. For the competing approaches, we evaluate different significance thresholds and present the best results.

\subsection{Independence Testing}

In this experiment, we illustrate the practical performance of \SCI, the $G^2$ test and conditional mutual information. In particular, we evaluate how well they can distinguish between true $d$-separation and false alarms.

To do so, we simulate dependencies as depicted in Fig.~\ref{fig:d_separation} and generated data under various samples sizes ($100$--$2 \, 500$) and noise settings ($0 \%$--$95 \%$). For each setup we generated $200$ data sets and assess the accuracy. In particular, we report the correct identifications of $F \independent T \mid D,E$ as the true positive rate and the false identifications $D \independent T \mid E,F$ or $E \independent T \mid D,F $ as false positive rate.

\begin{figure}[t]%
	\begin{minipage}[t]{.5\linewidth}
	\includegraphics[]{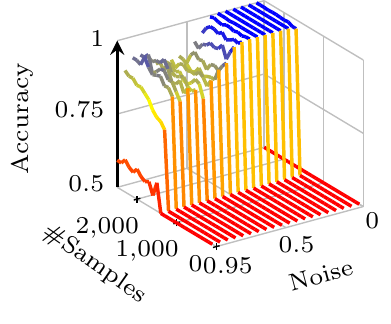}
	\end{minipage}%
	\begin{minipage}[t]{.5\linewidth}
	\includegraphics[]{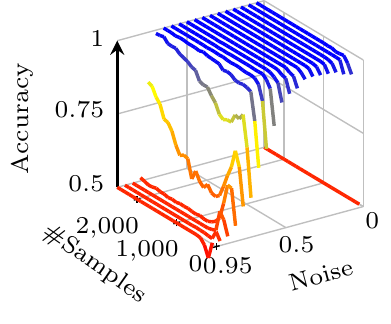}
	\end{minipage}%
	\caption{[Higher is better] Accuracy of $G^2$ (left) and \SCI (right) for varying samples sizes and additive noise percentages, where a noise level of $0.95$ refers to $95 \%$ additive noise.} 
	\label{fig:indep_comp}
\end{figure}

First, we focus on the results for $G^2$ and \SCI, which we show in Fig~\ref{fig:indep_comp}. \SCI performs with near to $100 \%$ accuracy for less than $70 \%$ noise and then starts to drop in the presence of more noise. At the level of $0 \%$ noise $D$ and $E$ can be fully explained by $F$ and therefore an accuracy of $50 \%$, i.e. all independences hold, is the correct choice. In comparison, the $G^2$ test marks everything as independent given less than $1 \, 500$ data points and starts to lose performance with more than $30\%$ noise. However, it is better in very high noise setups (more than $75\%$ noise), whereas it si questionable if the real signal can still be detected.

\begin{figure}[t]%
	\begin{minipage}[t]{.5\linewidth}
	\includegraphics[]{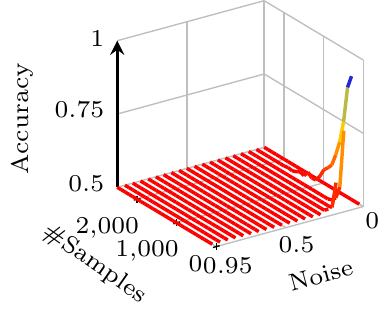}
	\end{minipage}%
	\begin{minipage}[t]{.5\linewidth}
	\includegraphics[]{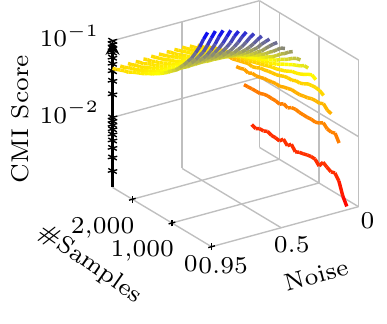}
	\end{minipage}%
	\caption{Accuracy of CMI (left) and the average value returned by CMI for the true independent case (right) for varying samples sizes and additive noise percentages. $I(F;T \mid D,E)$ is larger for small sample sizes.} 
	\label{fig:cmi_details}
\end{figure}

Next we consider the results for CMI. As theoretically discussed, we expect that finding a sensible cut-off value is impossible, since it depends on the size of the data, as well as the domain sizes of both the target and conditioning set. As shown in Fig.~\ref{fig:cmi_details}, CMI with zero as cut-off performs nearly random. In addition, we can clearly see that CMI is highly dependent on the sample size as well as on the amount of noise.

\subsection{Plug and Play with \SCI}

\begin{figure}[t]%
	\begin{minipage}[t]{.5\linewidth}
	\includegraphics[]{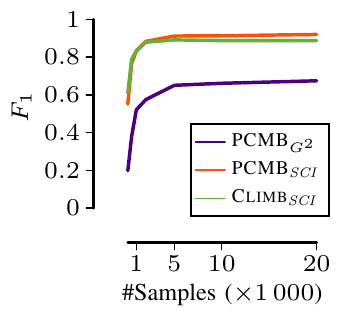}
	\end{minipage}%
	\begin{minipage}[t]{.5\linewidth}
	\includegraphics[]{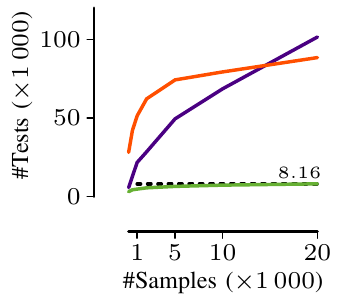}
	\end{minipage}%
	\caption{[Undirected MB] $F_1$ score and number of independence tests for $\pcmb_{G^2}$, $\pcmb_{\SCI}$ and $\ourmethod_{\SCI}$ on the \textit{Alarm} network for different sample sizes.}
	\label{fig:climb_comparison}
\end{figure}

To evaluate how well \SCI performs in practice, we plug it into both the \pcmb~\cite{pena:07:pcmb} and the \ipcmb~\cite{fu:08:ipcmb} algorithm. As the results for \ipcmb are similar to those for \pcmb, we skip them for conciseness. In particular, we compare the results of \pcmb using $G^2$ and \SCI to \ourmethod using \SCI and the \pcmb subroutine to find the parents and children. We refer to \pcmb with $G^2$ as $\pcmb_{G^2}$ and using \SCI as $\pcmb_{\SCI}$. To compare those variants, we generate data from the \textit{Alarm} network, where we generate for $100$ up to $20 \, 000$ samples each ten data sets and plot the average $F_1$ score as well as the number of performed independence tests in Fig.~\ref{fig:climb_comparison}.

As we can see, the $F_1$ score for \pcmb using $G^2$ reaches at most $67 \%$. In comparison, \pcmb using \SCI as well as \ourmethod obtain $F_1$ scores of more than $90 \%$. For both the precision is greater than $95 \%$ given at least $1 \, 000$ data points. Note that \ourmethod only uses the inferred children to search for the spouses and hence can at most be as good as $\pcmb_{\SCI}$. When we consider the runtime, we observe that \ourmethod has superior performance: Both the $\pcmb_{\SCI}$ and $\pcmb_{G^2}$ need more than 10 times as many tests than \ourmethod.

\subsection{Telling apart Parents and Children}

\begin{figure}[h]
	\includegraphics[]{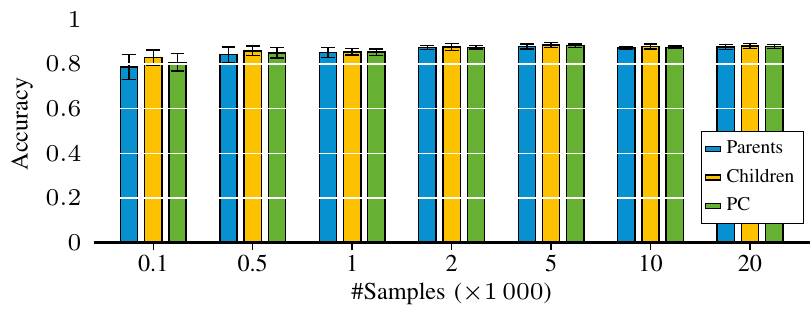}
	\caption{[Higher is better] Accuracy for telling apart parents and children given the parents and children for each node in the alarm network, given different sample sizes.}
	\label{fig:precision_pcmb_ipcmb}
\end{figure}

Next, we evaluate how well we can tell apart parents and children. We again generate synthetic data from the \textit{Alarm} network as above and average over the results. For each node in the network, we infer, given the true parents and children set, which are the parents and which the children using our score and plot the averaged accuracy in Fig.~\ref{fig:precision_pcmb_ipcmb}.

Given only $100$ samples, the accuracy is already around $80 \%$ and further increases to $88 \%$ given more data. 
In addition, the results show that there is no bias towards preferring either parents or children.

\subsection{Finding Causal Markov Blankets}

In the next experiment, we go one step further. Again, we consider generated data from the \textit{Alarm} network for different sample sizes. This time, however, we apply \ourmethod and compute the precision and recall for the directed edges. As a comparison, we infer the causal skeleton with stable PC~\cite{colombo:14:stablepc} using the $G^2$ test, let it orientate as many edges as possible and then extract the Markov blanket. 

\begin{figure}
	\begin{minipage}[t]{.5\linewidth}
	\includegraphics[]{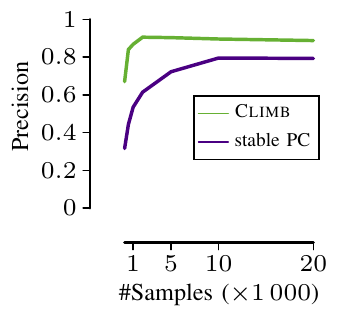}
	\end{minipage}%
	\begin{minipage}[t]{.5\linewidth}
	\includegraphics[]{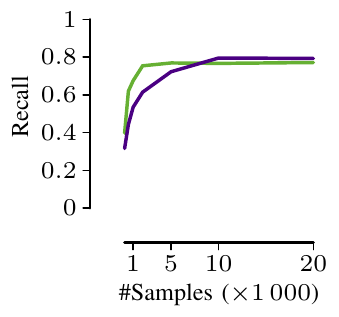}
	\end{minipage}%
	\caption{[Directed MB] Precision and recall of the \ourmethod and the PC algorithm for determining the causal MB on the \textit{Alarm} network for different sample sizes.}
	\label{fig:climb_pc}
\end{figure}

We plot the results in Fig.~\ref{fig:climb_pc}. We see that the precision of \ourmethod reaches up to 90\% and is always better than the \PCalgo algorithm, which reaches at most $79 \%$, for up to $10 \, 000$ data points. In terms of recall, \ourmethod is better than the PC algorithm, however, at $20 \, 000$ data points, they are about equal. Since \ourmethod is, as far as we know, the first method to extract the causal Markov blanket directly from the data, we can not have a fair comparison of runtimes. Given $20 \, 000$ data points, \ourmethod needs on average $221$ independence tests per node. 

\subsection{Causal Discovery}

Last but not least, we evaluate the use of \SCI, respectively \ourmethod, for causal discovery. First, we show that \SCI significantly improves the $F_1$ score over the directed edges of stable PC for small sample sizes compared to the $G^2$ test. Second, we apply the \ourmethod edge orientation procedure as post processing on top of the \fges and stable \PCalgo, and show how it improves their precision and recall on the directed edges. 

\paragraph{\SCI for PC}

First, we evaluate stable PC~\cite{colombo:14:stablepc}, using the standard $G^2$ test with $\alpha=0.01$ and \SCI on the \textit{Alarm} network with sample sizes between $100$ and $20 \, 000$. We generate ten data sets for each sample size and calculate the average $F_1$ score for stable PC using $G^2$ (stable $\PCalgo_{G^2}$) and and our independence test (stable $\PCalgo_{\SCI}$). To calculate the $F_1$ score, we use the precision and recall on the directed edges, which means that only if an edge is present with the correct orientation, we count it as a true positive.

\begin{figure}[h]
	\includegraphics[]{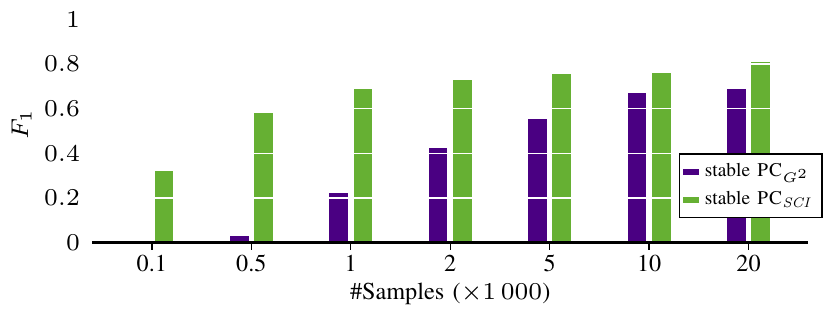}
	\caption{[Higher is better] $F_1$ score on directed edges for stable PC using $G^2$ and $SCI$ on the \textit{Alarm} network given different sample sizes.}
	\label{fig:pc_small}
\end{figure}

We show the results in Fig.~\ref{fig:pc_small}. Stable $\PCalgo_{\SCI}$ has a much better performance than stable $\PCalgo_{G^2}$, given only few samples. When the sample size approaches $20 \, 000$, stable $\PCalgo_{\SCI}$ still has an advantage of $\sim12 \%$ over $\PCalgo_{G^2}$.

\paragraph{Climb-based FGES and PC}

To show that our \ourmethod-based edge orientation scheme improves not only constraint based algorithms, but also score based methods, we apply it on top of \fges and stable \PCalgo, and correspond to the enhanced versions as $\fges_{\ourmethod}$ and $\PCalgo_{\ourmethod}$.
In Table~\ref{tab:causal_discovery} we record the average precision, recall and $F_1$ score. The tested networks networks \textit{Alarm}, \textit{Hailfinder}, \textit{Hepar2}, \textit{Win95pts} and \textit{Andes}, have between 37 and 223 nodes and can be found in the Bayesian Network Repository.\!\footnote{http://www.bnlearn.com/bnrepository/} We generate ten random data sets with $20 \, 000$ samples for each network and average over the results.

Applying \ourmethod to the undirected edges clearly improves both the results of \PCalgo and \fges. In all cases, except \fges on the \textit{Hailfinder} network, the precision and recall of the enhanced method are more than one standard deviation and up to $\sim 25 \%$ better than the original results.

To test the significance of these results, we apply the exact one-sided Wilcoxon rank-sum test~\cite{marx:16:wmw} on the precision and recall. As result, the enhanced versions significantly improve the precision and recall with all $p$-values $< 0.0018$.

\begin{table}[]
\caption{Performance of stable PC, FGES and their enhanced versions using \ourmethod-based edge orientation on $20 \, 000$ samples.}
\centering
\begin{tabular}{p{3em}p{1.8em}llll}
\toprule
\textbf{Dataset}&\textbf{Nodes}&\textbf{Approach}&\textbf{Precision}&\textbf{Recall}&\textbf{$F_1$} \\ \hline 
Alarm&37		&$\PCalgo$&$74.1 \pm 1.5$&$64.6 \pm 2.1$&$69.0$ \\
&			&$\PCalgo_{\ourmethod}$&\boldmath{$99.0 \pm 1.3$}&\boldmath{$86.3 \pm 1.7$}&\boldmath{$92.2$} \\
&			&$\fges$&$73.9 \pm 9.7$&$77.1 \pm 8.9$&$75.5$ \\ 
&			&$\fges_{\ourmethod}$&\boldmath{$85.3 \pm 5.2$}&\boldmath{$89.1 \pm 3.8$}&\boldmath{$87.1$} \\ \hline
Hailfinder&56	&$\PCalgo$&$79.1 \pm 1.6$&$42.4 \pm 0.0$&$55.2$ \\
&			&$\PCalgo_{\ourmethod}$&\boldmath{$92.4 \pm 1.8$}&\boldmath{$49.5 \pm 0.7$}&\boldmath{$64.5$} \\
&			&$\fges$&$82.2 \pm 4.7$&$82.6 \pm 3.9$&$82.4$ \\ 
&			&$\fges_{\ourmethod}$&\boldmath{$82.6 \pm 4.8$}&\boldmath{$82.9 \pm 3.9$}&\boldmath{$82.7$} \\ \hline
Hepar2&70	&$\PCalgo$&$62.1 \pm 2.8$&$34.0 \pm 2.0$&$43.9$ \\
&			&$\PCalgo_{\ourmethod}$&\boldmath{$66.1 \pm 3.1$}&\boldmath{$36.2 \pm 2.2$}&\boldmath{$46.8$} \\
&			&$\fges$&$85.2 \pm 2.5$&$61.9 \pm 1.7$&$71.7$ \\ 
&			&$\fges_{\ourmethod}$&\boldmath{$89.1 \pm 2.5$}&\boldmath{$64.7 \pm 1.8$}&\boldmath{$75.0$} \\ \hline
Win95pts&76	&$\PCalgo$&$83.9 \pm 1.0$&$59.7 \pm 1.5$&$69.8$ \\
&			&$\PCalgo_{\ourmethod}$&\boldmath{$94.7 \pm 0.7$}&\boldmath{$67.4 \pm 1.1$}&\boldmath{$78.7$} \\
&			&$\fges$&$82.4 \pm 3.7$&$78.1 \pm 3.9$&$80.2$ \\ 
&			&$\fges_{\ourmethod}$&\boldmath{$87.1 \pm 3.8$}&\boldmath{$82.6 \pm 3.9$}&\boldmath{$84.8$} \\ \hline
Andes&223	&$\PCalgo$&$94.3 \pm 0.9$&$79.8 \pm 0.8$&$86.4$ \\
&			&$\PCalgo_{\ourmethod}$&\boldmath{$97.0 \pm 0.8$}&\boldmath{$82.1 \pm 0.7$}&\boldmath{$89.0$} \\
&			&$\fges$&$95.7 \pm 0.5$&$86.0 \pm 1.0$&$90.6$ \\ 
&			&$\fges_{\ourmethod}$&\boldmath{$98.7 \pm 0.5$}&\boldmath{$88.7 \pm 1.0$}&\boldmath{$93.5$} \\
\bottomrule
\end{tabular}
\label{tab:causal_discovery}
\end{table}
\section{Conclusion}
\label{sec:conclusion}

This work includes three key contributions: \SCI, a conditional independence test for discrete data, the \ourmethod algorithm to mine the causal Markov blanket and the edge orienting scheme based on \ourmethod for causal discovery. Through thorough empirical evaluation we showed that each of those contributions strongly improves local and global causal discovery on observational data.

Moreover, we showed that \SCI converges to the true conditional mutual information. In contrast to CMI, it does not require any cut-off value and is more robust in the presence of noise. In particular, incorporating \SCI in either common Markov blanket discovery algorithms, or the PC algorithm leads to much better results than using the standard state of the art $G^2$ test---especially for small sample sizes. Further, we proposed \ourmethod to efficiently find causal Markov blankets in large networks. By applying our edge orientation scheme based on \ourmethod on top of common causal discovery algorithms, we can improve their precision and recall on the edge directions. 

For future work, we want to consider sparsification, by e.g. removing candidates that do not have a significant association to a target node. One possible way of doing this could be to formulate a significance test based on the no-hypercompression inequality~\cite{grunwald:07:book,marx:17:slope}. Last but not least, we want to develop fast approximations to find and orient the parents and children of hub nodes, to extend the applicability of our method.

\section*{Acknowledgment}
Alexander Marx is supported by the International Max Planck Research School for Computer Science (IMPRS-CS). Both authors are
supported by the Cluster of Excellence on ``Multimodal Computing and Interaction'' within the Excellence Initiative of the German Federal Government.

\bibliographystyle{IEEEtranS}
\bibliography{bib/abbrev,bib/bib-jilles,bib/bib-paper,bib/bib-alex}

\providecommand{\noopsort}[1]{}
\begin{thebibliography}{10}
\providecommand{\url}[1]{#1}
\csname url@samestyle\endcsname
\providecommand{\newblock}{\relax}
\providecommand{\bibinfo}[2]{#2}
\providecommand{\BIBentrySTDinterwordspacing}{\spaceskip=0pt\relax}
\providecommand{\BIBentryALTinterwordstretchfactor}{4}
\providecommand{\BIBentryALTinterwordspacing}{\spaceskip=\fontdimen2\font plus
\BIBentryALTinterwordstretchfactor\fontdimen3\font minus
  \fontdimen4\font\relax}
\providecommand{\BIBforeignlanguage}[2]{{%
\expandafter\ifx\csname l@#1\endcsname\relax
\typeout{** WARNING: IEEEtranS.bst: No hyphenation pattern has been}%
\typeout{** loaded for the language `#1'. Using the pattern for}%
\typeout{** the default language instead.}%
\else
\language=\csname l@#1\endcsname
\fi
#2}}
\providecommand{\BIBdecl}{\relax}
\BIBdecl

\bibitem{aliferis:10:hiton:overview}
C.~Aliferis, A.~Statnikov, I.~Tsamardinos, S.~Mani, and X.~Koutsoukos, ``{Local
  Causal and Markov Blanket Induction for Causal Discovery and Feature
  Selection for Classification Part I: Algorithms and Empirical Evaluation},''
  \emph{JMLR}, vol.~11, pp. 171--234, 2010.

\bibitem{aliferis:03:hiton}
C.~F. Aliferis, I.~Tsamardinos, and A.~Statnikov, ``Hiton: a novel markov
  blanket algorithm for optimal variable selection,'' in \emph{AMIA Annu Symp
  Proc}, vol. 2003.\hskip 1em plus 0.5em minus 0.4em\relax American Medical
  Informatics Association, 2003, pp. 21--25.

\bibitem{chaitin:75:algindepcond}
G.~J. Chaitin, ``A theory of program size formally identical to information
  theory,'' \emph{J.\ ACM}, vol.~22, no.~3, pp. 329--340, 1975.

\bibitem{chickering:02:ges}
D.~M. Chickering, ``Optimal structure identification with greedy search,''
  \emph{JMLR}, vol.~3, no. Nov, pp. 507--554, 2002.

\bibitem{colombo:14:stablepc}
D.~Colombo and M.~H. Maathuis, ``Order-independent constraint-based causal
  structure learning,'' \emph{JMLR}, vol.~15, no.~1, pp. 3741--3782, 2014.

\bibitem{cover:06:elements}
T.~M. Cover and J.~A. Thomas, \emph{Elements of Information Theory}.\hskip 1em
  plus 0.5em minus 0.4em\relax Wiley, 2006.

\bibitem{fu:10:review:mb:featureselection}
S.~Fu and M.~C. Desmarais, ``{Markov Blanket Based Feature Selection: a Review
  of Past Decade},'' \emph{WCE}, vol.~I, pp. 321--328, 2010.

\bibitem{fu:08:ipcmb}
------, ``Fast markov blanket discovery algorithm via local learning within
  single pass,'' in \emph{CSCSI}.\hskip 1em plus 0.5em minus 0.4em\relax
  Springer, 2008, pp. 96--107.

\bibitem{glymour:99:computation}
C.~N. Glymour and G.~F. Cooper, \emph{Computation, causation, and
  discovery}.\hskip 1em plus 0.5em minus 0.4em\relax Aaai Press, 1999.

\bibitem{grunwald:07:book}
P.~Gr\"{u}nwald, \emph{The Minimum Description Length Principle}.\hskip 1em
  plus 0.5em minus 0.4em\relax MIT Press, 2007.

\bibitem{janzing:10:algomarkov}
D.~Janzing and B.~Sch{\"o}lkopf, ``Causal inference using the algorithmic
  markov condition,'' \emph{IEEE TIT}, vol.~56, no.~10, pp. 5168--5194, 2010.

\bibitem{kolmogorov:65:information}
A.~Kolmogorov, ``Three approaches to the quantitative definition of
  information,'' \emph{Problemy Peredachi Informatsii}, vol.~1, no.~1, pp.
  3--11, 1965.

\bibitem{kontkanen:07:histo}
P.~Kontkanen and P.~Myllym\"aki, ``{MDL} histogram density estimation,'' in
  \emph{AISTATS}.\hskip 1em plus 0.5em minus 0.4em\relax JMLR, 2007, pp.
  219--226.

\bibitem{vitanyi:93:book}
M.~Li and P.~Vit\'{a}nyi, \emph{An Introduction to Kolmogorov Complexity and
  its Applications}.\hskip 1em plus 0.5em minus 0.4em\relax Springer, 1993.

\bibitem{mandros:17:noname}
P.~Mandros, M.~Boley, and J.~Vreeken, ``{Discovering Reliable Approximate
  Functional Dependencies},'' in \emph{KDD}.\hskip 1em plus 0.5em minus
  0.4em\relax ACM, 2017, pp. 355--364.

\bibitem{margaritis:00:gs}
D.~Margaritis and S.~Thrun, ``Bayesian network induction via local
  neighborhoods,'' in \emph{NIPS}, 2000, pp. 505--511.

\bibitem{marx:16:wmw}
A.~Marx, C.~Backes, E.~Meese, H.-P. Lenhof, and A.~Keller, ``{EDISON-WMW: Exact
  Dynamic Programming Solution of the Wilcoxon-Mann-Whitney Test},''
  \emph{Genomics, Proteomics {\&} Bioinformatics}, 2016.

\bibitem{marx:17:slope}
A.~Marx and J.~Vreeken, ``{Telling Cause from Effect using MDL-based Local and
  Global Regression},'' in \emph{ICDM}.\hskip 1em plus 0.5em minus 0.4em\relax
  IEEE, 2017, pp. 307--316.

\bibitem{mononen:08:sub-lin-stoch-comp}
T.~Mononen and P.~Myllym{\"{a}}ki, ``Computing the multinomial stochastic
  complexity in sub-linear time,'' in \emph{PGM}, 2008, pp. 209--216.

\bibitem{pearl:88:firstmb}
J.~Pearl, \emph{Probabilistic reasoning in intelligent systems: Networks of
  plausible inference}.\hskip 1em plus 0.5em minus 0.4em\relax Morgan Kaufmann,
  1988.

\bibitem{pearl:09:book}
------, \emph{Causality: Models, Reasoning and Inference}, 2nd~ed.\hskip 1em
  plus 0.5em minus 0.4em\relax New York, NY, USA: Cambridge University Press,
  2009.

\bibitem{pena:07:pcmb}
J.~M. Pe{\~{n}}a, R.~Nilsson, J.~Bj{\"{o}}rkegren, and J.~Tegn{\'{e}}r,
  ``{Towards scalable and data efficient learning of Markov boundaries},''
  \emph{Int J Approx Reason}, vol.~45, no.~2, pp. 211--232, 2007.

\bibitem{ramsey:17:fges}
J.~Ramsey, M.~Glymour, R.~Sanchez-Romero, and C.~Glymour, ``A million variables
  and more: the fast greedy equivalence search algorithm for learning
  high-dimensional graphical causal models, with an application to functional
  magnetic resonance images,'' \emph{Int J Data Sci Anal}, vol.~3, no.~2, pp.
  121--129, 2017.

\bibitem{rissanen:78:mdl}
J.~Rissanen, ``Modeling by shortest data description,'' \emph{Automatica},
  vol.~14, no.~1, pp. 465--471, 1978.

\bibitem{rissanen:86:stochastic}
------, ``Stochastic complexity and modeling,'' \emph{Annals Stat.}, pp.
  1080--1100, 1986.

\bibitem{schluter:14:survey:mb}
F.~Schl{\"u}ter, ``A survey on independence-based markov networks learning,''
  \emph{Artif.\ Intell.\ Rev.}, pp. 1--25, 2014.

\bibitem{shtarkov:87:universal}
Y.~M. Shtarkov, ``Universal sequential coding of single messages,''
  \emph{Problemy Peredachi Informatsii}, vol.~23, no.~3, pp. 3--17, 1987.

\bibitem{silander:08:nml:bayesnet}
T.~Silander, T.~Roos, P.~Kontkanen, and P.~Myllym{\"{a}}ki, ``{Factorized
  Normalized Maximum Likelihood Criterion for Learning Bayesian Network
  Structures},'' \emph{PGM}, pp. 257--264, 2008.

\bibitem{spirtes:00:book}
P.~Spirtes, C.~Glymour, and R.~Scheines, \emph{Causation, Prediction, and
  Search}.\hskip 1em plus 0.5em minus 0.4em\relax MIT press, 2000.

\bibitem{visweswaran:14:mbcounting}
S.~Visweswaran and G.~F. Cooper, ``Counting markov blanket structures,''
  \emph{arXiv preprint arXiv:1407.2483}, 2014.

\bibitem{zhang:10:iamb:lambda}
Y.~Zhang, Z.~Zhang, K.~Liu, and G.~Qian, ``{An improved IAMB algorithm for
  Markov blanket discovery},'' \emph{Journal of Computers}, vol.~5, no.~11, pp.
  1755--1761, 2010.

\bibitem{zhu:14:fast:ipcmb}
X.~Zhu and Y.~Yang, ``A fast markov blanket discovery algorithm,'' in
  \emph{ICSESS}.\hskip 1em plus 0.5em minus 0.4em\relax IEEE, 2014, pp.
  318--322.

\end{thebibliography}

\newpage
\clearpage

\section{Appendix}
\label{sec:appendix}

\subsection{Proof of log-concavity of the regret term}
\label{app:concave}

\begin{IEEEproof}[Proof of Lemma~\ref{lemma:log:concave}]
To improve the readability of this proof, we write $\regret_L^n$ as shorthand for $\regret_{\models_L}^n$ of a random variable with a domain size of $L$.

Since $n$ is an integer, each $\regret_L^n > 0$ and $\regret_L^0 = 1$, we can prove Lemma~\ref{lemma:log:concave}, by showing that the fraction $\regret_L^n / \regret_L^{n-1}$ is decreasing for $n \ge 1$, when $n$ increases.

We know from Mononen and Myllym{\"{a}}ki~\cite{mononen:08:sub-lin-stoch-comp} that $\regret_L^n$ can be written as the sum 
\begin{equation}
\label{eq:regretlin}
\regret_L^n = \sum_{k=0}^n m(k,n) = \sum_{k=0}^n \frac{n\fallingfactorial{k}(L-1)\risingfactorial{k}}{n^kk!}  \; ,
\end{equation}
where $x\fallingfactorial{k}$ represent falling factorials and $x\risingfactorial{k}$ rising factorials. Further, they show that for fixed $n$ we can write $m(k,n)$ as
\begin{equation}
\label{eq:regret:rekursion:k}
m(k,n) = m(k-1,n) \frac{(n-k+1)(k+L-2)}{nk} \; ,
\end{equation}
where $m(0,n)$ is equal to $1$. It is easy to see that from $n=1$ to $n=2$ the fraction $\regret_L^n / \regret_L^{n-1}$ decreases, as $\regret_L^0 = 1$, $\regret_L^1 = L$ and $\regret_L^2 = L + L(L-1)/2$. In the following, we will show the general case. We rewrite the fraction as follows.
\begin{align}
\frac{\regret_L^n}{\regret_L^{n-1}} &= \frac{\sum_{k=0}^n m(k,n)}{\sum_{k=0}^{n-1} m(k,n-1)} \\
&= \frac{\sum_{k=0}^{n-1} m(k,n)}{\sum_{k=0}^{n-1} m(k,n-1)} + \frac{m(n,n)}{ \sum_{k=0}^{n-1} m(k,n-1)} \label{eq:begin:proof}
\end{align}
Next, we will show that both parts of the sum in Eq.~\refeq{eq:begin:proof} are decreasing when $n$ increases. We start with the left part, which we rewrite to
\begin{align}
&\frac{\sum_{k=0}^{n-1} m(k,n-1) + \sum_{k=0}^{n-1} \left( m(k,n)- m(k,n-1) \right) }{\sum_{k=0}^{n-1} m(k,n-1)} \\
= &1+ \frac{\sum_{k=0}^{n-1} \frac{(L-1)\risingfactorial{k}}{k!} \left( \frac{n\fallingfactorial{k}}{n^k} - \frac{(n-1)\fallingfactorial{k}}{(n-1)^k} \right)}{\sum_{k=0}^{n-1} m(k,n-1)} \; . \label{eq:step2}
\end{align}
When $n$ increases, each term of the sum in the numerator in Eq.~\refeq{eq:step2} decreases, while each element of the sum in the denominator increases. Hence, the whole term is decreasing. In the next step, we show that the right term in Eq.~\refeq{eq:begin:proof} also decreases when $n$ increases. It holds that
\[
\frac{m(n,n)}{ \sum_{k=0}^{n-1} m(k,n-1)} \ge \frac{m(n,n)}{m(n-1,n-1)} \; .
\]
Using Eq.~\refeq{eq:regret:rekursion:k} we can reformulate the term as follows.
\begin{align}
&\frac{\frac{n+L-2}{n^2} m(n-1,n)}{m(n-1,n-1)} \\
= &\frac{n+L-2}{n^2} \left( 1 + \frac{m(n-1,n) - m(n-1,n-1)}{m(n-1,n-1)} \right)
\end{align}
After rewriting, it is easy to see that also the second term of Eq.~\refeq{eq:begin:proof} is decreasing and hence we can conclude the proof.
\end{IEEEproof}

\subsection{\SCI fulfills the Zero-Baseline Property}
\label{app:zero-baseline}

Given a random variable $X$ and a set of random variables $\mathbf{Y}$, where each $Y \in \mathbf{Y}$ is jointly independent of $X$. An independence measure fulfills the \textit{zero-baseline property}, if it remains zero, or indicates no association, independent of the sample size or the complexity of $\mathbf{Y}$~\cite{mandros:17:noname}. 

\begin{figure}[t]%
	\centering
	\includegraphics[]{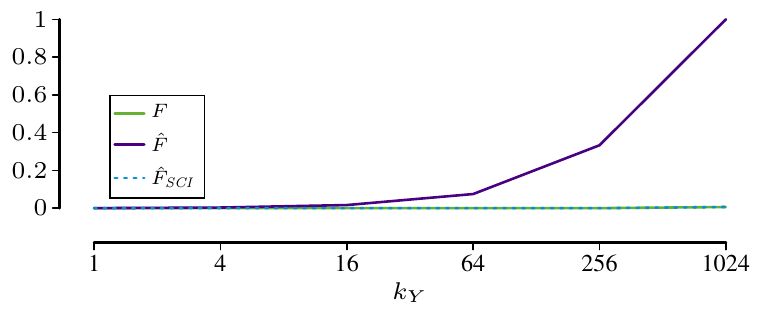}
	\caption{[Zero-Baseline] $\hat{F}$, $F$ and $\hat{F}_{\SCI}$ on $X \independent Y$, for different domain sizes $k_Y$ of $Y$. $F$ is the ideal score, $\hat{F}$ uses empirical estimates, while $\hat{F}_{\SCI}$ uses stochastic complexity. $\hat{F}$ identifies spurious associations for larger $k_Y$, whereas $\hat{F}_{\SCI}$ correctly does not pick up any signal.}
	\label{fig:zero_baseline}
\end{figure}

To illustrate that \SCI fulfills the zero-baseline property, while CMI does not, we consider their behaviour when the conditioning set is empty. We generate $X$ and $Y$ independently, setting their domain sizes respectively to $k_X=4$ while increasing $k_Y$ from $4^0$ to $4^5$. For a score between zero and one we divide $I$ by $H(X)$ and write $\hat{F}(X;Y) = I(X;Y \mid \emptyset) / H(X)$. We instantiate $F$ using the true entropy, $\hat{F}$ using empirical estimates of $H(\cdot)$, and $\hat{F}_{\SCI}$ using \SCI. For each setup between $X$ and $Y$ we generate $100$ data sets with $1 \, 000$ samples, average over the results and plot them in Fig.~\ref{fig:zero_baseline}. 
$\hat{F}_{\SCI}$ correctly identifies independence, whereas $\hat{F}$ almost immediately is non-zero, and quickly rises up to $1$, identifying a functional relationship (!) instead of independence.

\end{document}